\documentclass[]{trbunofficial}
\usepackage{graphicx}

\usepackage[hidelinks]{hyperref}

\usepackage{subcaption} 
\AuthorHeaders{Badu-Marfo and Farooq}
\title{Dynamic Campus Origin-Destination Mobility Prediction using Graph Convolutional Neural Network on WiFi Logs}

\author{%
  \textbf{Godwin Badu-Marfo}\\
  PostDoctoral Researcher \\
  Laboratory of Innovations in Transportation (LiTrans)\\
  Toronto Metropolitan University \\
  Toronto, Canada \\
  Email: gbmarfo@torontomu.ca\\
  \hfill\break
  \textbf{Bilal Farooq}\\
  Associate Professor\\
  Laboratory of Innovations in Transportation (LiTrans) \\
  Toronto Metropolitan University \\
  Toronto, Canada \\
  Email: bilal.farooq@torontomu.ca
  \hfill\break%
}

\usepackage{caption}

\usepackage{tabularx}
\usepackage{multirow}

\begin{document}
\maketitle

\section{Abstract}

We present an integrated graph-based neural networks architecture for predicting campus buildings occupancy and inter-buildings movement at dynamic temporal resolution that learns traffic flow patterns from Wi-Fi logs combined with the usage schedules within the buildings. The relative traffic flows are directly estimated from the WiFi data without assuming the occupant behaviour or preferences while maintaining individual privacy. We formulate the problem as a data-driven graph structure represented by a set of nodes (representing buildings), connected through a route of edges or links using a novel Graph Convolution plus LSTM Neural Network (GCLSTM) which has shown remarkable success in modelling complex patterns. We describe the formulation, model estimation, interpretability and examine the relative performance of our proposed model. We also present an illustrative architecture of the models and apply on a real-world WiFi logs collected at the Toronto Metropolitan University campus. The results of the experiments show that the integrated GCLSTM models significantly outperform traditional pedestrian flow estimators like the Multi Layer Perceptron (MLP) and Linear Regression.

\hfill\break%
\noindent\textit{Keywords}: Dynamic OD, deep learning, temporal occupancy, pedestrian flows, graph neural networks.
\newpage

\section{Introduction}

Pervasive smart devices having the capability of wireless connectivity systems ingest enormous data that offer the potential to unravel non-intrusive behavioural patterns on a mobile population which is harnessed through intelligent algorithms \cite{badu2019perspective}. Most urban cities are embarking on initiatives and policies to improve public safety, health and security through a city-wide deployment of intelligent devices and monitoring systems (CCTV, Speed Recorders, WiFi Hotspots, sensors) to observe the near real-time state of physical infrastructure like Transportation systems. For example, the City of Toronto has piloted a free Wi-Fi project trial to extend free internet to low-income neighbourhoods during the COVID-19 pandemic \cite{FreeWiFi}. Similarly, public crowded areas like Airports, Parks, University Campus and Transit Hubs provide WiFi hotspots to offer internet connectivity to its users. Passive data gathered from these pervasive systems are largely enormous, and has the potential of harnessing knowledge discovery and intelligence for optimal decision making. \citep{poucin2018activity} used WiFi connection history to mine valuation information about the usage of an open public space. Also, \citep{farooq2015ubiquitous} used a multi-sensor network (infrared, RGB, WiFi) to study pedestrian dynamics of activities in a large public festival in Montreal.

Understanding the mobility behavioral patterns and predicting pedestrian activity play an important role in the optimal design of new infrastructure, such as university campuses, shopping centers, transit hubs, and in the daily operations of these infrastructure \cite{vo2020modeling}. Generally, mobility activities are spatially distributed over a temporal lapse and achieved through multiple modes (Cars, Bicycles, Bus, Train, Walk) but commonly associated with vehicles, walking is an important mode of travel especially for the first-mile and last-mile of each trip. Studying the pedestrian flow by walking within facilities is of importance in demand forecasting, occupancy prediction, resource management and towards new initiatives such as smart cities. GPS data has predominantly been used in mobility behavioral studies coupled with supplementary datasets such as land use, socio-demographic surveys to improve prediction accuracy and also to integrate the spatial population dynamics. However, for active surveys using GPS, the data owner will need to consent to share their data which could lead to privacy leakages and sampling bias \cite{badu2019perturbation}. Location-aware technologies such as WiFi, Cellular networks, Bluetooth and infrared exhibit the potential to passively collect data through detecting the presence of user's within a space and time \cite{badu2019perspective}. Typically, Wi-Fi devices store data on connection history of devices for service management and improvement. These Wi-Fi logs contain timestamp of connection, Access Point (AP) identifier and a unique user device identifier in the form of a Media Access Control (MAC) address. APs provide WiFi access having a connection to internet services and are spatially distributed covering large areas (e.g. airports, campus, shopping malls).

Earlier research works have delved into traffic information predictions including traffic flow, speed  and road occupancy using machine learning and time series models such as Hidden Markov Models (HMM), AutoRegressive Integrated Moving Average (ARIMA) and Deep Learning models like Recurrent Neural Networks and  Long Short-Term Memory (LSTM). While these learning algorithms are capable of solving traffic prediction problems, they lack the ability to integrate the spatio-temporal features and natural graph structure attributes of traffic data. To solve this, the Graph Convolution Neural Networks (GGNN) have shown promising successes in updating weights to a graph structure and maintains the spatial correlations between feature nodes connected through weighted edges.   In this regard, the GCNN is gaining adoption for solving traffic problems due to its inherent ability to represent problems in a graph structure. We advance the current state of research by proposing a Graph Convolution plus LSTM neural network (GCLSTM) architecture to predict traffic flow of pedestrians commuting within facilities of a university campus. We propose a methodology for pedestrian flow prediction that integrate spatio-temporal features including  building occupancy, entry and exit count of flow within buildings.

The rest of the article is structured as follows: a detailed review of current literature on the use of ubiquitous networks for traffic flow analysis and graph neural networks architecture.  This leads us to a section on the methodology, results and analysis related to short-term pedestrian flows between campus buildings. In the end, we discuss our conclusions, limitation and possible applications.

\section{Literature review}
\label{sec:background}

The domain of communication-enabled networks and pervasive devices (cellular networks, Global Positioning Systems (GPS) or WiFi networks) \cite{badu2019perspective} have grown enormously both in academic research and industrial adoption in the last decade. Transportation studies have used data sourced from these ubiquitous systems to study the mobility patterns and behavioral dynamics of the population, and it's interaction with limited infrastructure resources. This interest has birthed knowledge discovery in network optimization, urban modeling and transportation policy. In the past, transportation studies used data gathered from origin-destination (OD) surveys and census data conducted through interviews, questionnaires, and telephone interviews for a sample population to draw conclusions on the mobility patterns and behaviour of the entire population. These approaches lead to large sampling bias, and high cost of data collection that requires experts like researchers and experiment participants to be engaged. 

The progress in communication technologies and the drawbacks of traditional travel survey methods birthed automated travel surveys based on GPS technology that is largely used in transportation research. The potential of GPS surveys in gathering more accurate spatio-temporal characteristics of trips eventually made it a suitable replacement to paper-form surveys. GPS tracking data has been used for mode of transport detection (\cite{rezaie2017semi, zheng2008understanding, dabiri2018inferring}), trip segment detection \cite{patterson2016datamobile, zhou2016data}, trip activity detection \cite{stopher2009travel, ermagun2017real} and transit itinerary inference \cite{zahabi2017transit, wu2016travel}. Typically, the GPS requires data collection in outdoor spaces with cloud visibility and is ineffective in closed spaces like buildings or tunnels. Hence, the adoption of GPS data for building occupancy mobility is minimal because it results in low positional accuracy and noise.

To address the drawbacks of the GPS, Wireless-Fidelity (WiFi) data has gained traction in traffic intelligence applications, and used in problems including trajectory prediction \cite{wang2016trajectory}, activity recognition \cite{poucin2018activity}, mode choice detection \cite{kalatian2018mobility} and next location choice. WiFi Access Points are continuously deployed in major cities especially public areas that attract crowd for socio-economic purposes, to provide access to internet services at free or low charges. The Wi-Fi data is passively collected whenever a user connects to the network without the requirement of consent from the participant. The Wi-Fi logs present a location-aware fine-grained precision for the purpose of indoor applications like occupancy and traffic flow between buildings. 

Prediction of traffic flow is a critical component in the development and expansion of ITS, for the purpose of optimizing the traffic time and reducing traffic volumes. Parametric models have a fixed number of parameters for predicting traffic flow including the linear regression model, Kalman filter model and time series model (ARIMA). The autoregressive integrated moving average (ARIMA) model is widely used in traffic flow prediction \cite{williams2001multivariate, ghosh2009multivariate}. \citep{li2012prediction} used ARIMA models to predict the number of passengers at hot-spot pick-up locations using Taxi GPS traces. However, the accuracy of the ARIMA model is low for extreme value predictions and the model is not suitable for traffic conditions in the real world because could lead to peaks and fast fluctuations \cite{davis1990adaptive}. To address the shortcomings of the ARIMA model, the extended Kalman filter is adopted and performs well in short-term and real-time traffic predictions \cite{wang2007real, guo2014adaptive}. Non-parametric models like the k-nearest neighbor model \cite{zheng2014short}, the Bayesian network \cite{sun2006bayesian}, the support vector model \cite{wu2004travel} and the neural network model \cite{chan2011neural} has been used for traffic flow forecasting. Short-term traffic flow predictions with DL Models have gained popularity in the field of traffic prediction \cite{liu2021scientometric}, including Convolution Neural Networks (CNN) \cite{zhang2019short}, Recurrent Neural Networks (RNN) \cite{osipov2020urban} and Long Short-Term Memory (LSTM) \cite{tian2018lstm}.

Most of these deep learning models extract the local patterns of data, but do not efficiently integrate spatial and hierarchical dependencies existing in data features. The Graph Neural Networks (GNNs) have the capacity to represent the spatial correlation of graph-based data, making it suitable for spatial-temporal features and dynamic correlations of traffic data. \citep{yu2017spatio} proposed a gated graph convolution network for traffic prediction but does not consider the dynamic spatial-temporal correlations of traffic data.

\section{METHODOLOGY}

In this section, we discuss the detailed methodology used to develop the GCLSTM architecture for campus-level pedestrian flow. First of all, we formulate the campus pedestrian flow problem with dependencies on the occupancy state of the Origin-Destination (O-D) buildings. In addition, the data cleaning, processing and aggregation assumptions performed on the WiFi data are discussed in depth. Secondly, we discuss the graph structure with representation for mobility trajectories, and the model parameter tuning used for the implementation of GCLSTM architecture in this work. At the concluding part of this section, we discuss the benchmark metrics adopted to evaluate the GCLSTM model. 

\begin{figure}[ht]
	\centering
	\includegraphics[width=\textwidth]{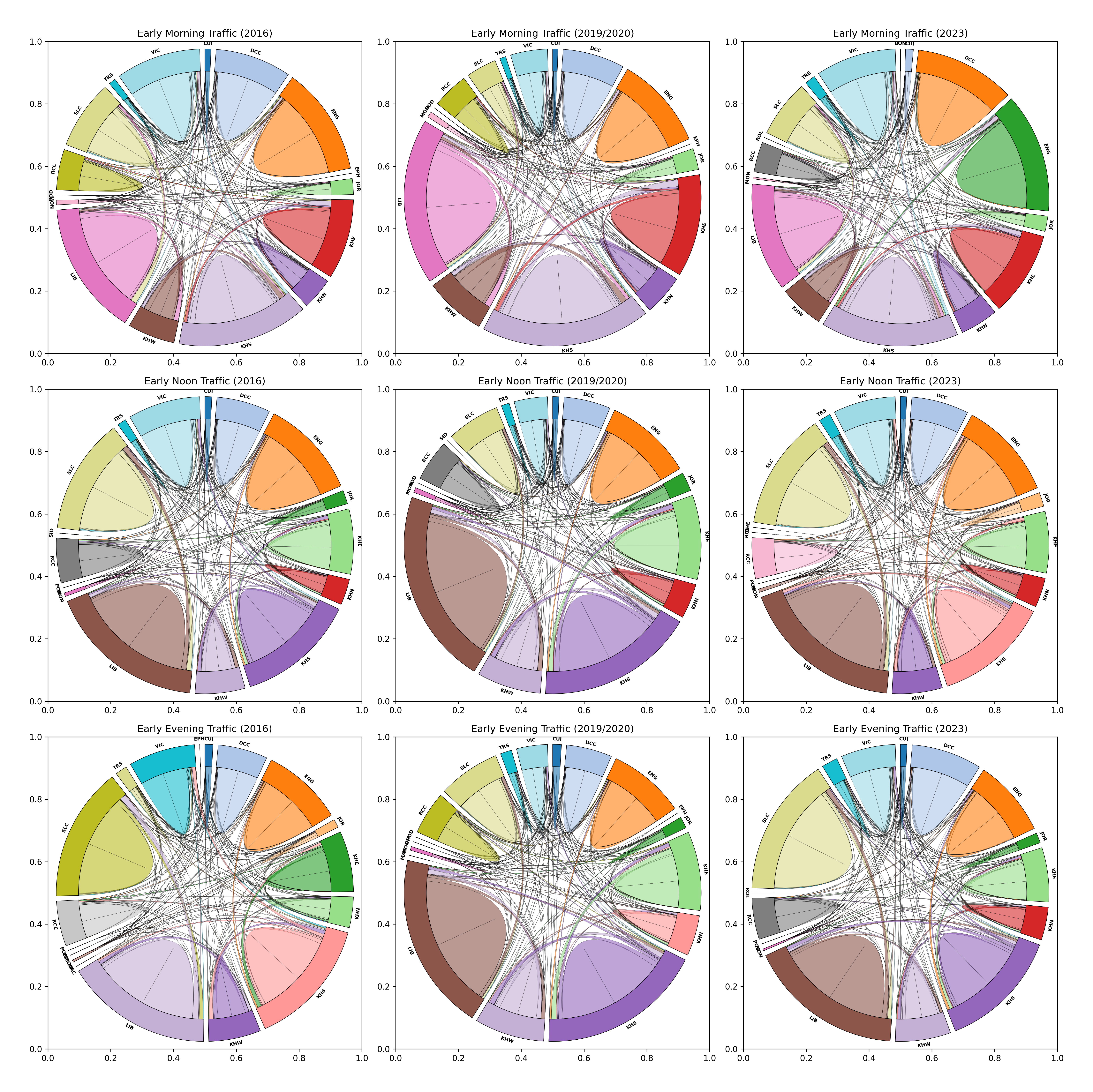}
	\caption{Time Periods Comparison of Pedestrian Traffic Patterns}
	\label{fig:pedestrain_traffic}
\end{figure}

\begin{figure}[ht]
	\centering
	\includegraphics[width=\textwidth]{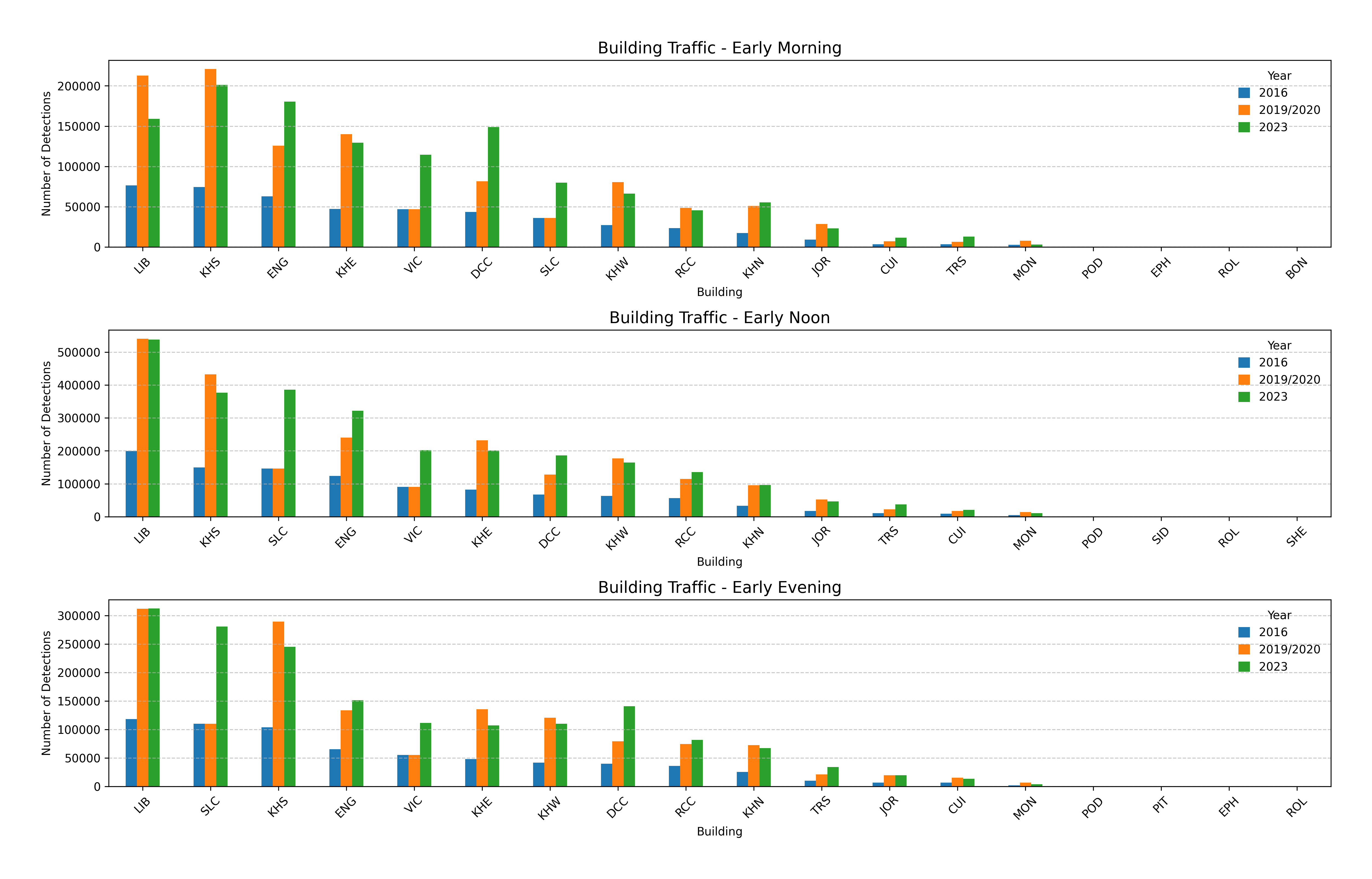}
	\caption{Building Traffic Counts observed at Peak Periods}
	\label{fig:building_traffic}
\end{figure}

\subsection{Problem Definition}

We assume individuals (pedestrians) commute largely by walking within campus facilities to undertake their daily socio-economic activities such as attend class lectures (education), perform professional roles (work) or engage in social activities with friends (leisure) on the campus premises. We leverage on redacted WiFi logs to learn the mobility patterns and temporal occupancy within buildings on the university campus. The WiFi logs are collected in four (4) weeks over three years (2016, 2019, 2023). 

The WiFi devices record the unique MAC address and timestamp of the individual's devices at any time a network connection is established. Using the WiFi logs, the mobility dynamics of pedestrians can be studied and their complete trip trajectories recreated. We aggregate building occupancy by the count of unique MAC ids observed in a building for peak time periods as shown in Figure \ref{fig:pedestrain_traffic}. Similarly, pedestrian flow is estimated as the count of trip chains observed between buildings marked by the log of MAC Ids who sequential logs are recorded in the WIFi records of multiple buildings shown in Figure \ref{fig:building_traffic}. For example, if a mac id is logged at 8:00am in CUI building and consequently logged at LIB building at 8:15am, then we assume a trip chain suggesting the building of first timestamp as the trip origin and the building of the last timestamp as the trip destination. 

The goal of this work is to estimate pedestrian flow between buildings using the observed mobility dynamics within buildings at previous time steps formulated as;

\begin{linenomath}
    \begin{equation} \label{eq1}
        PF_{t+1} = BO_{t}
    \end{equation}
\end{linenomath}\\

where $PF_{t+1}$ is the predicted passenger flow at time \textit{t + 1} and $BO_{t}$ represents the mobility dynamic features observed in building at time \textit{t}. The mobility dynamic features are aggregate counts of extracted patterns from WiFi logs using rule-based assumptions. We assume that the MAC Id of a device in a log denotes the presence of the device owner in the building at the logged timestamp. A sequence of time-ordered logs by the same MAC ID suggests the trip trajectories taken by the individual within the campus premises. We also assume the first log of the individual (MAC Id) denotes the first entry of the individual to the campus and the last log as the exit of the individual from the campus. We use these assumptions to build the data points for the model inputs. To apply these assumptions, we will highlight the case study, the Wi-Fi data and pre-processing of the Wi-Fi data in the next section.

\subsection{Case Study and Dataset}

\begin{figure}[ht]
	\centering
	\includegraphics[width=\textwidth]{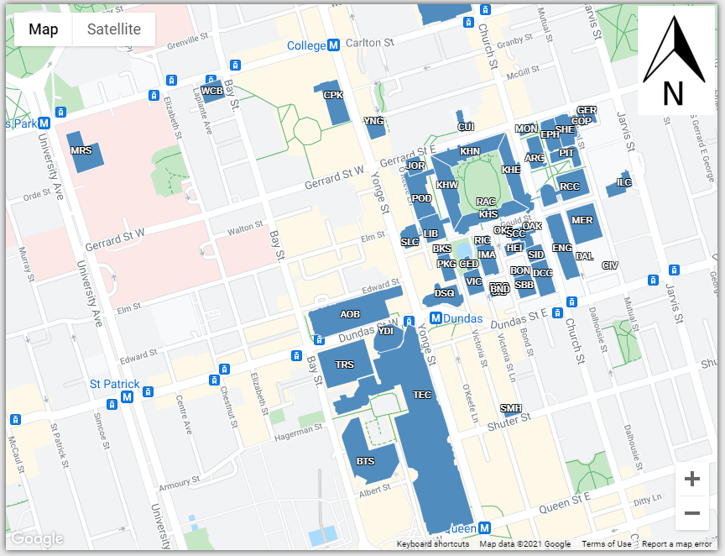}
	\caption{Map of the study region: Toronto Metropolitan University Campus, Toronto}
	\label{fig:map_tmu}
\end{figure}

In this study, we collected WiFi logs from Wireless Access Points or Routers installed in the campus buildings of the Toronto Metropolitan University shown in the Figure \ref{fig:map_tmu}. These logs were recorded when users of the campus connected their devices to the university network for internet services. Data logs were collected from a total of twenty-two (22) WiFi Access points within a four (4) weeks duration in 2016, 2019 and 2023. The logs consist of the devices that are authenticated and connected to the network, while the devices in exploration mode are excluded. The advantage of this filter is that we have access to the unique MAC ID of the device throughout the connection. The downside of this approach is that the resulting logs represent the pattern of a very large sample rather than the population. This point can be addressed by developing similar weights as developed in \citep{farooq2015ubiquitous}. The key features extracted from the data logs are described in the table \ref{table:wifi_data} below.

\begin{table}[ht]
    \centering
    \begin{tabular}{|l|l|l|}
    \hline
        Field & Description & Source \\ \hline
        mac\_id & MAC address of user & WiFi \\ \hline
        wifi\_id & SSID of the access points & WiFi \\ \hline
        log\_date & Timestamp of the connection logs & WiFi \\ \hline
        building\_id & Campus building Id & Class Schedule \\ \hline
        enrolment\_no & Total Enrolment Number & Class Schedule \\ \hline
    \end{tabular}
    \caption{Field schema of WiFi logs}
    \label{table:wifi_data}
\end{table}

\begin{figure}[ht]
	\centering
	\includegraphics[width=\textwidth]{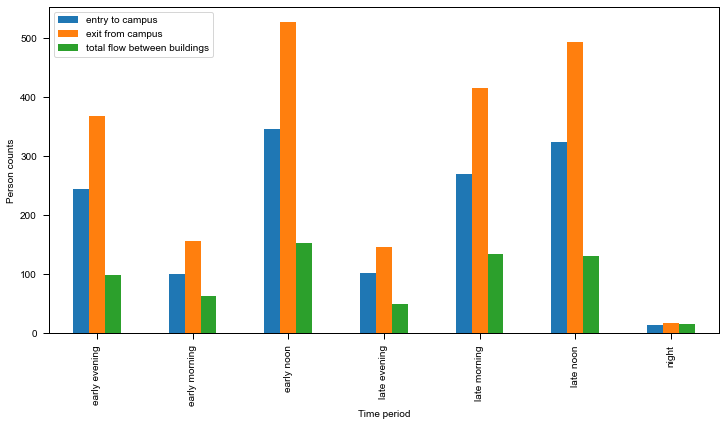}
	\caption{Traffic flow count between buildings observed at TMU}
	\label{fig:traffic_tmu}
\end{figure}

\subsection{Pre-Processing of WiFi logs}

\begin{figure}[ht]
	\centering
	\includegraphics[width=\textwidth]{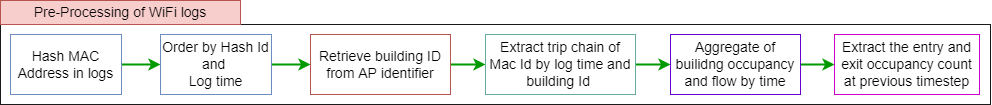}
	\caption{Node representation of campus building}
	\label{fig:processing}
\end{figure}

In this section, we discuss the step approach used to process and extract the important latent features from the raw Wi-Fi data as show in Figure \ref{fig:processing}. These features served as input to the Graph model. 

The WiFi data contains tuples of Media Access Control (MAC) address, ``MAC Id" of the connecting device, the Service Set Identifier ``SSID" of the wireless access point, and the timestamp of the log. Generally, the MAC address is a privacy sensitive information that could pose risk to the device owner if the sequence of logs are gained by an adversary. To solve this, we first redact the MAC Ids by hashing, to produce a random series of letters and numbers to replace the MAC Ids. Using this technique, attackers are incapable to reverse the hashed IDs to retrieve the MAC address thus privacy preservation is achieved.

Subsequent to the Hashing of MAC IDs, the new hash IDs are ordered by the logged timestamps to map the trip sequences of devices such that the location geometry is defined by the location of the associated Wi-Fi Ids. In step 3, the SSID of the WiFi APs are scaled to building they are installed in such that every SSID is replaced with its building ID. Building level trip chains are generated for each unique device marked by the hash Id. This is followed by the count aggregation of trip OD for each building segmented at specified time intervals (15min, 30min, 60min). Building occupancy is computed as the count of unique hash Ids present in a building at a time interval while pedestrian flow is computed as the count of hash Ids that moved to other buildings at a time interval shown in Figure \ref{fig:traffic_tmu}.

Finally, we model the mobility dynamics for each building by calculating the temporal entry and exit count for each building. These features suggest the number of people who entered the campus for the first time or exited the campus at a time interval. We compute this by observing the timestamp of the first and last log of every device for a specified interval then aggregate all first logs as entries or last logs as exist for each building. 

We summarize the input features that explain the variances of mobility dynamics of buildings in Table \ref{table:mobility_features}. These features are used as inputs features to the proposed model. We discuss the detail the architecture of the model in the following section.

\begin{table}[ht]
\centering
\begin{tabular}{|l|l|}
\hline
\textbf{feature} & \textbf{short description}                                           \\ \hline
$l_{entry}$ & persons first logged in campus network upon entry into building \\ \hline
$l_{exit}$  & persons last logged in campus network upon exit from building   \\ \hline
$l_{orig}$  & persons whose trip originated from building                     \\ \hline
$l_{dest}$  & persons whose trip completed in building                        \\ \hline
$l_{time}$  & timestamp of log observations                                   \\ \hline
\end{tabular}
\caption{Table showing features of mobility dynamics}
\label{table:mobility_features}
\end{table}

\subsection{Graph Convolution Neural Network for Pedestrian Flow Learning}

In this work, we adapt the Graph Convolution Neural Network \citep{kipf2016semi} to develop a regressive architecture for pedestrian flow estimation. The GCNN employs the concept of graph structures to the traditional convolution neural network approach, each pixel of an image applies filter matrix to its neighboring pixel to achieve an averaged feature map of pixel neighbors. In graphs, nodes with similarity are more likely to be connected to each other than dissimilar ones, a concept called ``network homophily'' \cite{zhu2020beyond}. Information about each node is stored in a feature vector, and latent information on the cross correlations between nodes is achieved by aggregating a node's features with that of its neighbors. This operation of neighborhood aggregation is synonymous to the convolution of images.

\citep{kipf2016semi} proposed a Graph Convolution Neural Network (GCNN) with a weighted average operation to balance the uneven spread of feature vectors on isolated nodes with less neighbors. This is implemented by a normalize operation that assigns bigger weights to feature vectors from nodes with few neighbors, formulated as:

\begin{linenomath}
    \begin{equation}
    \label{eq2}
        h_{i} = \sum_{j \in {N}_{i}} \frac{1}{\sqrt{{deg}(i)} \sqrt{{deg}(j)}} W_{i} x_{j}
    \end{equation}
\end{linenomath}\\

where $W_i$ is the unique weight vector for each node, $x_j$ is the feature vector of a node, and $deg$ is the degree of node.

\begin{figure}[ht]
	\centering
	\includegraphics[width=\textwidth]{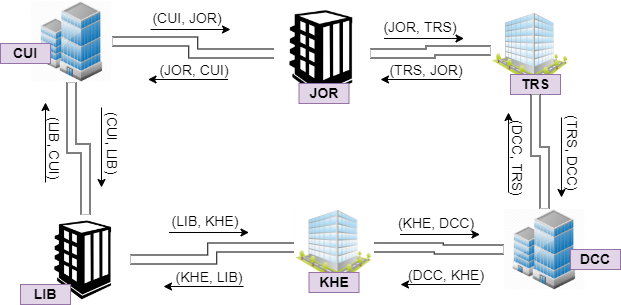}
	\caption{Node representation of campus building}
	\label{fig:building_node}
\end{figure}

We formulate the pedestrian flow as a node regression problem that takes point-based features of the mobility dynamics within buildings at previous timestamps as input, using GCNN as the base. In order to implement the model, we first discuss the pedestrian flow problem in the context of graph structure representation made up of nodes and edges as shown in Figure \ref{fig:building_node}.

\subsubsection{Node representation}

In this work, a set of campus buildings $B = (b_1, b_2, b_3, b_4, ..., b_n)$ representing the unique campus buildings that are spatially distributed on the premises. To estimate the pedestrian flow between buildings, a unique combination of the buildings are created such that each building is paired with other buildings such as $[(b_1, b_2), (b_1, b_3), (b_2, b_3) .... (b_i, b_j)]$. Each combination pair represents the direction of pedestrian flow. For instance, a pair $(b_m, b_n)$ represents the aggregation of pedestrian flow from building $m$ to the building $n$. These building pair are used to represent the nodes or vertices $v$ of the graph as shown in Figure \ref{fig:building_node}. In addition, each node takes in node features $v_k$ made of the building occupancy features at previous timestamp $t-1$ including the occupancy of paired buildings, the persons logged first on entry into the campus at $t-1$ and the persons last logged out or exit from the campus at $t-1$. Hence, each node is represented by a vector of aggregate features as shown in Table \ref{table:mobility_features}.

\begin{figure}[ht]
	\centering
	\includegraphics[width=0.5\textwidth]{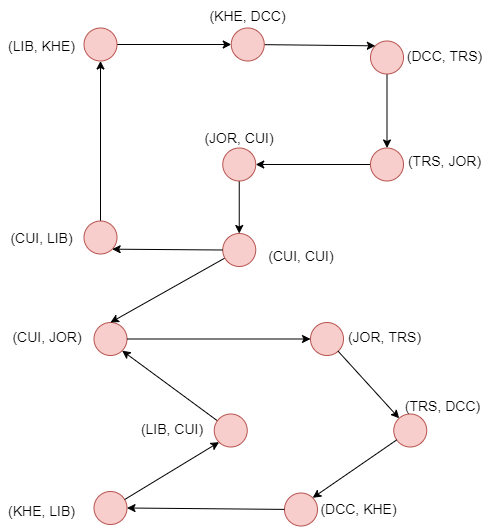}
	\caption{Graph nodes representation of campus building}
	\label{fig:building_graph}
\end{figure}

\subsubsection{Edge representation}

\begin{figure}[ht]
	\centering
	\includegraphics[width=0.5\textwidth]{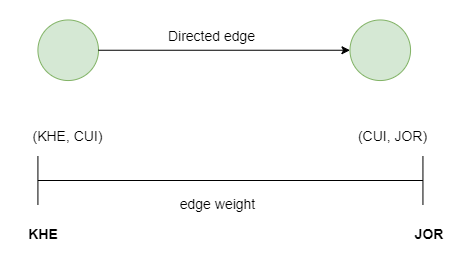}
	\caption{Graph edge representation of building links}
	\label{fig:building_edge}
\end{figure}

Edges represent the link to connect nodes of the graph. A pair of nodes $(b_i, b_j)$ and $(b_j,b_k)$ are assumed to be connected on a directed edge if the last element of the first pair $b_j$ is equal to the first element of the second pair $b_j$ as shown in Figure \ref{fig:building_graph} . We compute the edge weight to be the euclidean distance of the geographic coordinates marked by the first element of the first pair to the last element of the second pair. As shown in Figure \ref{fig:building_edge}, the edge weight is calculated by the euclidean distance of \textit{KHE} being the first element of the first pair \textit{(KHE, CUI)}, and \textit{JOR} the last element of the next pair \textit{(CUI, JOR)}. With the defined edges, an adjacency matrix is created to represent the graph connectivity. 

\subsubsection{Model configuration}

\begin{figure}[ht]
	\centering
	\includegraphics[width=0.8\textwidth]{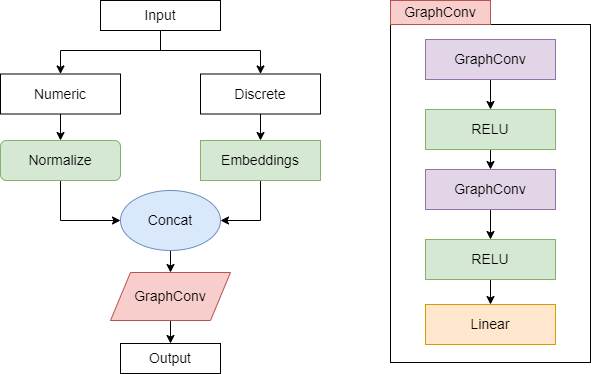}
	\caption{Model Architecture of Graph Convolutions}
	\label{fig:graph_model}
\end{figure}

With the numerical graph representation discussed above, we adapt the Graph Convolution Neural Network (GCNN) architecture to model passenger flow in a campus mobility setting. The model shown in Figure \ref{fig:graph_model} is composed of two(2) graph convolutional layers and a linear output layer. A rectified linear unit ``relu" activation functions is applied to the model layers to account for non-linearities that could exist. The model compiles using the ``adam" optimizer and a loss function of mean squared error. The model inputs are the graph node vector matrices that are scaled and normalized for the numeric features, and vector embeddings for the discrete features. Similarly, the output features are scaled and normalized between a range of negative one (-1) and positive (+1).

\section{Evaluation and results}

In this section, we evaluate the performance of the predicted pedestrian flow of the graph model. Also, we evaluate this performance metrics on other learning algorithms; the Multi Layer Perceptron (MLP)  and a standard linear regression. For all models, the same feature vectors are used at inputs to the model. The model was developed in PyTorch and experimented on a Windows 10 PC with Intel Core i7-2600 (8 Cores) and G-Force GTX 950.

We quantify the prediction accuracy between the observed and predicted passenger flow with the Standard Root Mean Square Error (SRMSE), the model fitness using a measure of the Pearson Correlation Coefficient (corr) and the coefficient of determination (R$^2$). The standardized root mean squared error is defined by:

\begin{linenomath}
    \begin{equation} \label{eq4}
    SRMSE(\hat{\pi},\pi)=\frac{RMSE(\hat{\pi},\pi)}{\bar{\pi}}=\frac{\sqrt{\sum _i \cdots \sum _j(\hat{\pi} _i... _j -  \pi _i... _j)^2/N _b}}{\sum _i ... \sum _j \pi _{i...j}/N _b}
    \end{equation}
\end{linenomath}\\

where \textit{N$_b$} is the total number of samples;  $\hat{\pi}$ and $\pi$ is the predicted and observed passenger flow respectively.

\subsection{Goodness of Fit Tests Analysis on varying time intervals}

\begin{figure}
\centering
  \begin{subfigure}[t]{.4\textwidth}
    \centering
    \includegraphics[width=\linewidth]{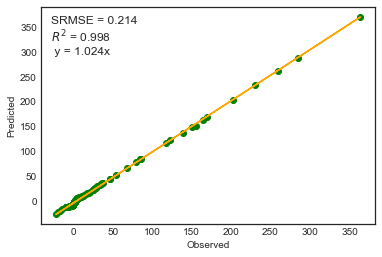}
    \caption{Pedestrian flow prediction at intervals of 15 minutes}
  \end{subfigure}
  \hfill
  \begin{subfigure}[t]{.4\textwidth}
    \centering
    \includegraphics[width=\linewidth]{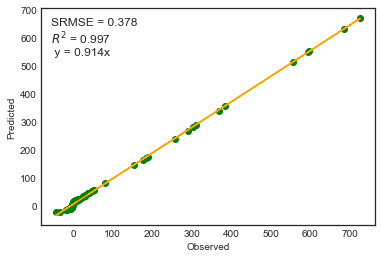}
    \caption{Pedestrian flow prediction at intervals of 30 minutes}
  \end{subfigure}

  \medskip

  \begin{subfigure}[t]{.4\textwidth}
    \centering
    \includegraphics[width=\linewidth]{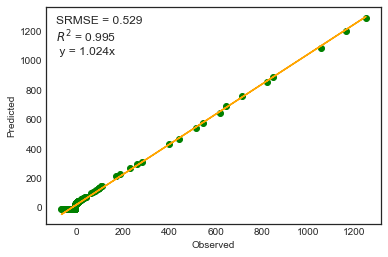}
    \caption{Pedestrian flow prediction at intervals of 60 minutes}
  \end{subfigure}
  
  \caption{Goodness of fit tests for Pedestrian flow predictions at specified time intervals}
  \label{fig:flow_prediction}
\end{figure}

To evaluate on the performance of the implemented model, we infer model prediction on test data set. We observe the RMSE and the adjusted R-squared values of the model by a goodness of fit test on the ground truth values against the model predicted values of passenger flow. This test is evaluated on the target variable of passenger flows using varying time intervals of 15, 30 and 60 minutes. The model performed considerably well for all time intervals. In Figure \ref{fig:flow_prediction}, the RMSE values of 0.214, 0.378, 0.529 were observed for passenger flow predictions at time interval of 15, 30 and 60 minutes. These average aggregates over all prediction outputs suggest the performance of the model. The RMSE values depict a monotonic increment in residual error when predictions are tested on longer time intervals. However, the model shows better performance at pedestrian flow counts on smaller intervals. From this results, we observe that the Graph Convolution Neural Network can capture the mobility dynamics in a campus setting without any background knowledge on pedestrian behaviour and we are also able to achieve a better accuracy with a graph representation of the pedestrian flow propagation.

\subsection{Comparative analysis to Machine Learning algorithms  }

\begin{figure}[ht]
  \begin{subfigure}[t]{.4\textwidth}
    \centering
    \includegraphics[width=\linewidth]{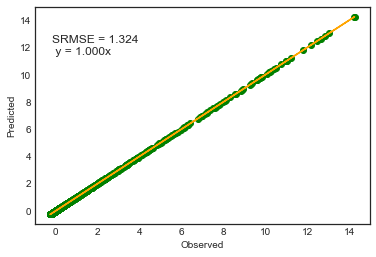}
    \caption{Pedestrian flow prediction using Linear Regression}
  \end{subfigure}
  \hfill
  \begin{subfigure}[t]{.4\textwidth}
    \centering
    \includegraphics[width=\linewidth]{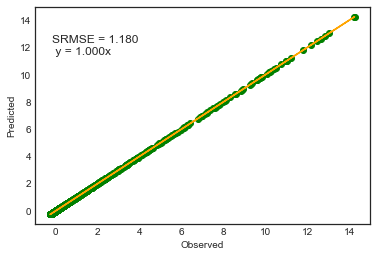}
    \caption{Pedestrian flow prediction using Multi Layer Perceptron}
  \end{subfigure}
  
  \caption{Goodness of fit tests comparisons with ML algorithms}
  \label{fig:ml_flow_prediction}
\end{figure}

We evaluate the proposed GCNN model against traditional machine learning algorithms to predict pedestrian flow using the same mobility dynamic features as input (see Figure \ref{fig:ml_flow_prediction}). We experiment on Multi Layer Perceptron (MLP) and Linear Regression (LR). We observe RMSE values of 1.324 and 1.180  for relative predictions using LR and MLP respectively. Both models show higher residual errors suggesting lower prediction accuracy than the proposed GCNN model. The GCNN performs better due to its capacity to integrate connectivity weights by node neighbors.

\subsection{Ablation Analysis with Enrolment Data}
\begin{figure}[ht]
	\centering
	\includegraphics[width=0.5\textwidth]{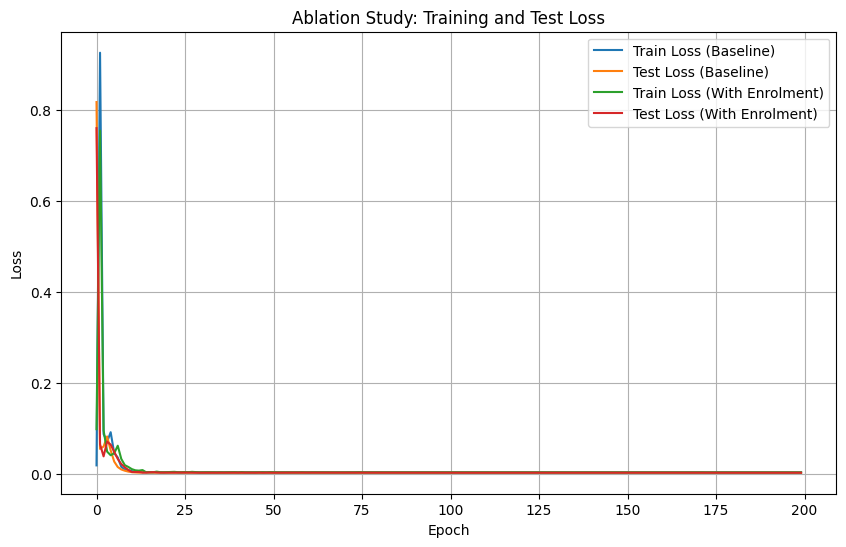}
	\caption{Ablation Study with/without enrolment data}
	\label{fig:ablation_study}
\end{figure}

To assess the effectiveness of the model, we perform ablation study with and without the the inclusion of course enrolment data as an additional feature for predicting inter-building flows. We assume buildings with class enrolments will express high mobility dynamics during the operating hours of the campus. In this analysis, we include the enrolment count as an additional node feature on the graph model. The performance of the analysis is shown in Figure \ref{fig:ablation_study}. With a baseline model with no enrolment data, the loss curves typically decrease and stabilize to suggest the model is learning meaningful patterns from temporal building-to-building count data.  However, with the enrolment feature, the loss curves of the model with enrolment are consistently lower than the baseline model which suggests the enrolment data provides an improved predictive power. The significant reduction in test loss with enrolment data indicates that class sizes and scheduled activities are important drivers of pedestrian flows between buildings.

\section{Conclusion and Discussion}

We developed a Graph Convolution Neural Networks architecture for estimating pedestrian flow at the campus level. The proposed architecture is demonstrated on the WiFi AP data from the Toronto Metropolitan University for 2016, 2019, and 2023. We harnessed the spatial topology capability of the graph model, to define the problem in a graph structure thus achieve a weighted neighbour aggregation of input features based on connected building pair nodes, a problem that is deficient in traditional machine learning algorithms. We demonstrated the use of graph model for a node regression problem having inputs of weighted short-term occupancy and mobility dynamic features of buildings. The proposed model performed well in learning the optimal fit with reduced residual errors in predicting future pedestrian flows. 

This work assumes point-based temporal input features without considering the sequential continuity of mobility dynamics of pedestrian flow. While the GCNN gave satisfactory results in the prediction of point-based features, it lacks the capability to learn long term dependencies, a requirement for temporal learning. Newer graph architectures have evolved for temporal learning, we will further this research to model the long-term dependencies of pedestrian flow using graph-based time-series or sequential learning algorithms. In addition, we will extend this work towards a multi-output estimation of flow and occupancy features for future time intervals. Demonstration of spatial transferability by using the WiFi data from other campuses is another important future dimension.

\section*{AUTHOR CONTRIBUTION STATEMENT}
\noindent
The authors confirm contribution to the paper as follows:\\
Study conception and design: GBM, BF \\
Data preparation: GBM \\
Methodology and investigation: GBM, BF \\
Analysis and interpretation of results: GBM \\
Draft manuscript preparation: GBM, BF \\
Funding and supervision: BF \\
All authors reviewed the results and approved the final version of the manuscript.

\newpage
\bibliographystyle{trb}
\bibliography{trb_template}

\end{document}